*Original Article*

# YOLO-APD: Enhancing YOLOv8 for Robust Pedestrian Detection on Complex Road Geometries

Aquino Joctum[1], John Kandiri[2]

[1,2]Department of Computing and Information Science (CIS), Kenyatta University, Kenya.

[1]Corresponding Author : joctuhm@gmail.com

***Abstract*** - *Autonomous vehicle perception systems require robust pedestrian detection, particularly on geometrically complex roadways like Type-S curved surfaces, where standard RGB camera-based methods face limitations. This paper introduces YOLO-APD, a novel deep learning architecture enhancing the YOLOv8 framework specifically for this challenge. YOLO-APD integrates several key architectural modifications: a parameter-free SimAM attention mechanism, computationally efficient C3Ghost modules, a novel SimSPPF module for enhanced multi-scale feature pooling, the Mish activation function for improved optimization, and an Intelligent Gather & Distribute (IGD) module for superior feature fusion in the network's neck. The concept of leveraging vehicle steering dynamics for adaptive region-of-interest processing is also presented. Comprehensive evaluations on a custom CARLA dataset simulating complex scenarios demonstrate that YOLO-APD achieves state-of-the-art detection accuracy, reaching 77.7% mAP@0.5:0.95 and exceptional pedestrian recall exceeding 96%, significantly outperforming baseline models, including YOLOv8.*

*Furthermore, it maintains real-time processing capabilities at 100 FPS, showcasing a superior balance between accuracy and efficiency. Ablation studies validate the synergistic contribution of each integrated component. Evaluation on the KITTI dataset confirms the architecture's potential while highlighting the need for domain adaptation. This research advances the development of highly accurate, efficient, and adaptable perception systems based on cost-effective sensors, contributing to enhanced safety and reliability for autonomous navigation in challenging, less-structured driving environments.*

## 1. Introduction

The advancement of Advanced Driver Assistance Systems (ADAS) and Autonomous Vehicles (AVs) promises to revolutionize road safety and transportation efficiency [1]. A key component of realizing this potential is creating a strong perceptual system to analyse difficult and evolving surroundings. Within these systems, pedestrian detection is critical; it's the key to preventing accidents that claim around 1.3 million lives globally every year [2], [3]. Notwithstanding the advancements made by Deep Learning (DL) and Computer Vision (CV)[4], accurate and timely pedestrian detection, especially in unstructured, geometrically complex and unpredictable scenarios, is still a big challenge [5]. This is further exacerbated by high sensor costs, limited hardware, erratic human behaviour[6], [7].

For example, developing regions sometimes have unique infrastructural problems, such as poorly planned roads, rugged terrains, and unpredictable pedestrian traffic, with people often sharing space with automobiles [8]. Such conditions severely test conventional perception systems. A particularly challenging situation central to this study is detecting pedestrians and moving hazards on complicated road geometries, notably Type-S (serpent) roads with sharp, sometimes blind, curves as illustrated in Figure 1. Such geometries dynamically limit the field of vision, generating regular occlusions and significantly lowering the available reaction time to hazards such as pedestrians emerging suddenly from curves. Furthermore, high-profile incidents, such as the fatal accident involving an Uber AV unable to classify a jaywalking pedestrian [2] correctly, underscore the essential requirement for systems that can adapt and react effectively in real-time, especially when faced with unforeseen circumstances and small or occluded targets.

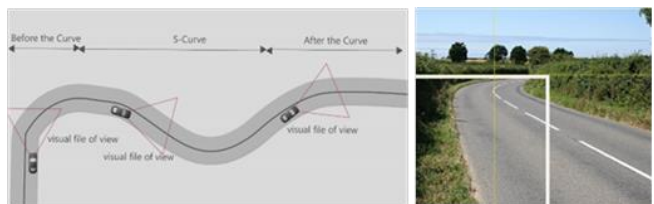

**Fig. 1 Type S Road and the difficulty of detecting hazards by AV**

State-of-the-art pedestrian detection systems frequently rely on expensive sensor suites combining LiDAR, RADAR, and multiple camera types [9], [10]. While multi-sensor fusion offers robustness [11], the associated cost and computational demands limit scalability and widespread adoption, making consumer-grade AVs increasingly impractical. As such, approaches leveraging cost-effective RGB cameras are gaining traction in academia and the field of Autonomous driving. However, RGB-based methods inherently contend with challenges such as adverse weather, occlusions, variable illumination (e.g., bright sun glare, nighttime), and the inherent unpredictability of human behaviour, all of which can impair algorithmic performance [12].

Despite the advancements in deep learning technologies and the exponential growth in object detection frameworks, a clear research gap persists. This gap lies not only in the development of a pedestrian detection system that simultaneously achieves high accuracy and real-time processing specifically for geometrically complex road environments but also in using only cost-effective RGB sensors, while also incorporating architectural innovations tailored to overcome issues like multi-scale feature degradation, inefficient feature fusion in deep networks, and the need for robust feature representation under challenging visibility. Existing models, including advanced YOLO iterations, may not fully optimize the balance between these specific requirements for such demanding scenarios without significant modification.

This paper introduces YOLO-APD (YOLO Adaptive Pedestrian Detection), a novel deep learning architecture to address this gap. Through targeted architectural changes that make it work better on complicated road shapes, YOLO-APD greatly improves the YOLOv8 framework. This work's primary problem is the insufficient accuracy and adaptability of current RGB-based detectors when faced with combined challenges of Type-S roads, potential occlusions and the need for split-second decision making. The primary contributions and novelty of this work are summarised as follows:

- Novel YOLO-APD Architecture: Based on the YOLOv8 benchmark, YOLO-APD introduces a unique integration of a SimSPPF module that improves multi-scale feature pooling over the standard SPPF module by incorporating Mish activation and SimAM attention. This is done to get back fine-grained details that might have been lost in aggressive pooling. Computationally efficient C3Ghost modules are an alternative to the standard C3 blocks to lower parameters and FLOPs while keeping feature representation, a strategy derived from GhostNet principles.
- A parameter-free SimAM attention mechanism is strategically added into the backbone to improve feature maps by highlighting important neuronal information that helps distinguish pedestrians in cluttered scenes, offering an edge over more complex, parametrised attention modules.
- A novel methodology leveraging vehicle steering dynamics to define an adaptive Region of Interest (ROI), significantly enhancing computational efficiency and focusing detection resources on trajectory-relevant areas, especially critical in complex road scenarios like Type-S curves. (This might be visually represented in Figure 2, which shows YOLO-APD implementation, if included.)
- The proposed YOLO-APD network can effectively extract multi-scale detail features in complex road and environmental conditions, with an exceptional recall reaching more than 96% with fewer false detections.

## 2. Related Work

The realization of a fully autonomous driving system is critically dependent on an autonomous vehicle's robust environmental perception, with pedestrian detection representing a paramount safety imperative [2]. To put the proposed YOLO-APD in the context of the larger trend of improvements in adaptive pedestrian detection systems, this study combines ideas from computer vision (CV), deep learning (DL), and adaptive systems (AS). Foundational CV research sets important rules for understanding visual scenes [13], [14]. DL methods allow learning hierarchical feature representations from the perceived raw visual data (Goodfellow et al., 2016). At the same time, AS frameworks improve the ability of systems to respond to changes in their environment [15].

Early pedestrian detection systems depended on hand-crafted features like Haar-like descriptors (Viola & Jones, 2001) and Histograms of Oriented Gradients (HOG) (Dalal & Triggs, 2005), which were often used with classifiers like Support Vector Machines (SVMs) to identify objects of interest. Despite their historical significance, these early models lacked robustness in the presence of occlusions, varying illumination, and intra-class variance. The rise of deep learning, especially Convolutional Neural Networks (CNNs), changed the way we think about how accurate and reliable object detection is [16]. R-CNN [17], Fast R-CNN [18], and Faster R-CNN [19] were the first two-stage detectors to use region-based classification with better accuracy. Still, they often had latency problems that made them less useful in real-time autonomous systems [20].

To mitigate inference delays being experienced with the existing traditional approaches, a single-stage class of object detectors emerged. They introduced a unique methodology that offered direct mappings from image pixels to bounding box predictions, accelerating speed and detection accuracy. Notable among these algorithms that promised a brighter prospect were SSD [21], [22] and the YOLO (You Only Look Once) family [23] for their remarkable performance. YOLOv1, the first entrant in the YOLO family, used a grid-based technique to make predictions. Later versions, such as YOLOv2 [24], YOLOv3 [25], and YOLOv4 [26], added anchor boxes, multi-scale pyramids, and deeper backbones. Since then, YOLOv5 [27], [28], YOLOv6 [29], YOLOv7 [30], YOLOv8 [24], [31], and YOLOv9 [32] have all incrementally improved the balance between speed and accuracy. They have done this by adding features like anchor-free designs and CSPNet-inspired backbones.

YOLOv8, the baseline algorithm for this study, show these innovations through an efficient, anchor-free architecture. However, even better performing models such as YOLOv8 can suffer performance degradation in scenes characterized by dense occlusions, small-scale or distant pedestrians, and visually cluttered environments [33], [34]. These challenges get worse in places with complicated road geometries, such as Type-S Road curves, where constrained visibility and dynamic field-of-view shifts strain detection reliability. Addressing these deficits constitutes a central motivation for YOLO-APD.

Recent architectural advances have introduced attention mechanisms, which emulate human visual prioritization to enhance feature focus. Squeeze-and-Excitation (SE) blocks [35], Efficient Channel Attention (ECA-Net) [36], Coordinate Attention [37], and SimAM [38], a parameter-free module incorporated in YOLO-APD, exemplify these approaches. Although attention modules have been integrated into YOLO variants [39], [40], they are typically optimized for general improvements rather than for the unique challenges posed by curved and occluded roadways.

Simultaneously, lightweight backbones such as GhostNet [41] are embedded into the C3Ghost module to preserve real-time efficiency. Multi-scale feature fusion, vital for robust object detection across spatial resolutions, is efficiently handled by optimised structures like PANet [42] and BiFPN [43]. YOLO-APD's neck design is influenced by Intelligent Gather-and-Distribute (IGD) principles, enhancing cross-scale feature interaction. In addition, the use of advanced activation functions—such as Mish [44]—and dynamic convolutions [45], [46] provides improved representational adaptability.

The contribution of YOLO-APD lies in its unique architectural enhancements of the YOLOv8 framework, which combines SimSPPF, C3Ghost, SimAM, an IGD-inspired neck module, and Mish activation function into a single improved algorithm. Using only monocular RGB sensors, this innovative assembly addresses the perceptual requirements of pedestrian detection in visually complex road geometries and low visibility conditions where current models frequently perform poorly.

The selection of sensor modality is a critical factor in determining detection quality in autonomous vehicles. Despite being widely available and moderately priced, RGB cameras are known to be sensitive to unfavourable weather conditions and inadequate lighting [12]. Although LiDAR and RADAR provide accurate depth and velocity information, they come at the expense of greater complexity and costs [9], [10]. On the other hand, infrared and thermal cameras provide better nighttime performance [47], [48]. Worth noting is that multimodal fusion raises system costs even though it provides robustness. Therefore, YOLO-APD concentrates on optimizing performance using inexpensive monocular RGB inputs. Research on autonomous vehicle perception typically falls into Line-of-Sight (LOS) and Non-Line-of-Sight (NLOS) techniques. Line-of-Sight (LOS) approaches for spotting pedestrians prove effective when sensors have an unobstructed view, yet their inherent limitations quickly surface. It's when pedestrians or other objects of interest are hidden from view that Non-Line-of-Sight (NLOS) techniques really become critical for reliable awareness [49].

Still, the reality is that LOS-reliant detection has largely taken centre stage in how autonomous vehicles perceive their surroundings. This strong preference for LOS isn't without consequence; it has markedly shaped the makeup and built-in biases of key public datasets like KITTI, Cityscapes [50], and nuScenes [51], which are commonly used in autonomous driving-related studies. These datasets, even Though comprehensive, it lacks sufficient diversity in extreme weather conditions, highly complex road geometries like Types S roads, besides relying on feature axis-aligned bounding boxes that don't fully capture occlusions. These hurdles motivate the approach taken in this study, which involves using the ARLA simulator [52] to generate a tailored dataset specifically designed to address these challenging scenarios.

Despite these substantial advancements in pedestrian detection, a persistent research gap remains. There is a need for models that can leverage inexpensive sensors to achieve high accuracy, recall, and real-time performance in urban areas with complex road geometries. The fact that vehicle kinematic data, like steering angle, is underutilized in visual models exacerbates this problem even further. Notably, existing networks often face performance issues with occlusions, scale differences, and visual clutter, especially in complex road geometries that pose a high risk to road safety.

YOLO-APD addresses this multifaceted problem by combining novel YOLOv8 architectural enhancements (including SimSPPF, C3Ghost, SimAM, and an IGD-inspired neck) with a dynamic ROI mechanism driven by steering input. This integrated approach, grounded in the theoretical frameworks of Computer Vision for image interpretation, Deep Learning for feature extraction, and Adaptive Systems for dynamic responsiveness, provides a single, useful framework for advanced pedestrian detection in the next generation of self-driving cars.

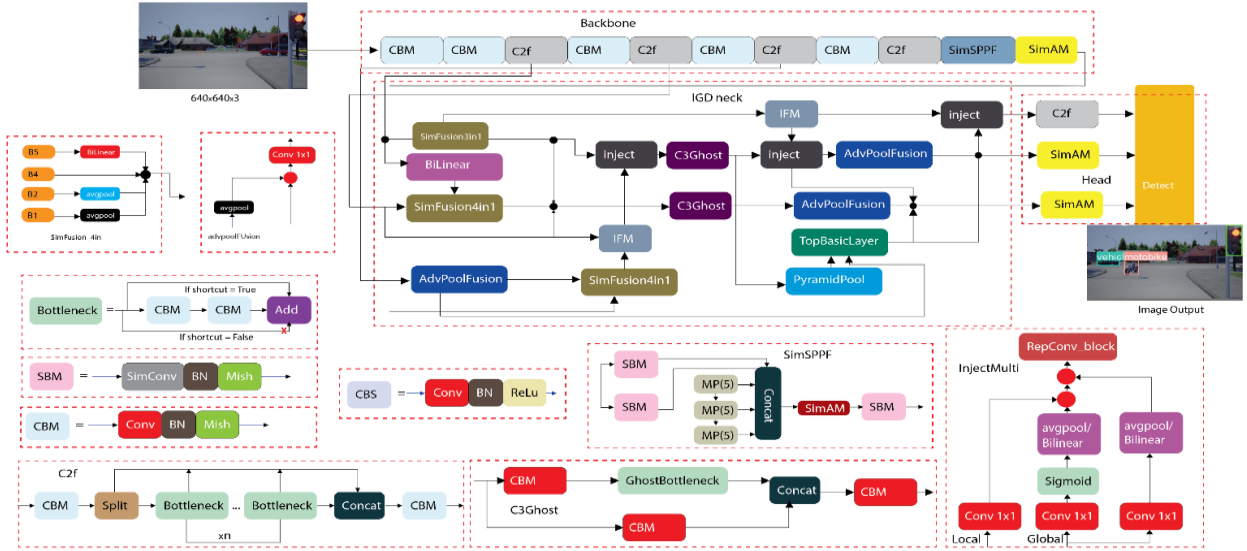

**Fig. 2 YOLO-APD Implementation**

## 3. Materials and Methods
### 3.1. Dataset Preparation
#### 3.1.1. CARLA Synthetic Dataset Generation
In this study, a dataset of 2015 images was generated using the CARLA simulator (v0.9.15) running on Unreal Engine 4, as it was difficult to use existing datasets, given their limitations and unsuitability for this study. Simulations utilized Town021-Town05 maps, selected for their diverse urban/suburban layouts, including road sections exhibiting sharp curves and elevation changes analogous to Type-S roads.

The dataset captured varied conditions: different times of day (daylight, twilight, night), weather (clear, rainy, foggy), and traffic scenarios with varying densities of dynamic vehicles and pedestrians Images (640x480 pixels) were captured from a forward-facing RGB recording camera mounted on an ego-vehicle operating in autopilot mode. 2015 images were obtained and pre-processed by cleaning the images by removing spaces and special characters.

Manual annotations were created for 10 classes: 'person', 'car', 'truck', 'bike', 'motorbike', 'traffic light green', 'traffic light red', 'traffic light orange', 'traffic sign 30', 'traffic sign 90'. The dataset was split into training (1753 images, 87%), validation (183 images, 9%), and test (79 images, 4%) sets.

#### 3.1.2. KITTI Benchmark Dataset
The widely recognized KITTI Object Detection Benchmark was used for evaluating the real-world generalization of the proposed model. To ensure conformity with the synthetic data classes, the study focused on the 'Person' and 'Car' categories from the standard validation split, using established evaluation protocols. Images were processed at a resolution of 640x640 pixels for training and validation.

The dataset was divided into a training set of 6732 images (90%) and a validation set of 749 images (10%).

### 3.2. YOLO-APD Network
#### 3.2.1. SimAM Attention Mechanism
A parameter-free attention module known as SimAM is introduced into the proposed detection network to enhance feature extraction for the proposed model. This module is inspired by studies of the mammalian nervous system that discovered that neurons of higher elevation have a domineering effect on peripheral neurons. Thus, this module (Figure 3) manipulates feature maps in CNN layers by generating 3D attention weights [53]. Different focus is given to the various areas of interest, depending on the significance of the information carried by neurons in those regions.

This causes the network to prioritize important characteristics within the target zones while improving the target's feature representation. Unlike conventional techniques, the module generates 3D weights without requiring extra subnetworks. This is enabled by directly calculating the weights using a particular energy function. The 3D weights for the present neuron can be inferred from the analytical solution specified by the function. All neurons of the same dimension are treated equally by the current attention module, which is worth mentioning. While considering the significance of both channel and spatial dimensions, this module assigns unique weight to each neuron, which is consistent with the properties of human Attention. The SimAM mechanism is also introduced into the backbone to verify its generalization in different tasks.

Finding neurons containing more significant information in visual neuroscience involves determining if target neurons are linearly separable from nearby neurons. Neurons carrying varying quantities of information fire in distinct ways. The formulation of the energy function can be minimized to achieve linear separability. The energy function is initially explained following the addition of the regularization term.

$$e_{t(w_1,b_t,y,x_i)} = \frac{1}{M-1}\sum_{i=1}^{M-1}(-1-(w_t x_i + b_t))^2 + (1-(w_t t + b_t))^2 + \lambda w_t^2 \quad (1)$$

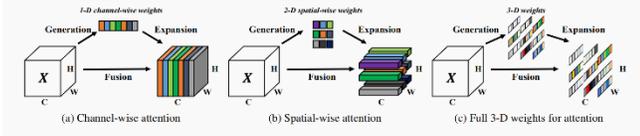

**Fig. 3 Comparison of different attention steps. Unlike typical attention mechanisms that generate 1D or 2D weights from features X before expanding them for channel (a) or spatial (b) attention, the proposed module directly computes 3D weights (c). The subfigures illustrate this: a consistent color represents a single scalar applied per channel, spatial location, or feature point. Adapted from [38]**

Where $t$ and $x_i$ Are the target and surrounding neurons of the input feature $X=R^{C \times H \times W}$ in a channel, $M=H \times W$ indicates how many neurons are present in that channel; and i denotes the exact loci of the spatial location. $wt$ and $bt$ are the weight and bias transforms [38]. Further, the specific forms of the analytical solutions of $w_t$ and $b_t$ are defined using Equations x and x. Where $\mu_t = \frac{1}{M-1}\sum_{i=1}^{M-1} x_i$ and $\sigma_t^2 = \frac{1}{M-1}\sum_{i=1}^{M-1}(x_i + \mu_t)^2$ Represent the mean and variance of each neuron except $t$, respectively.

It then follows that the minimum energy of the system can be calculated using the formula below:

$$e_t^* = \frac{4(\hat{\sigma}^2 + \lambda)}{(t-\hat{\mu})^2 + 2\hat{\sigma}^2 + 2\lambda} \quad (2)$$

Here, $\hat{\mu} = \frac{1}{M-1}\sum_{i=1}^{M-1} x_i$ and $\hat{\sigma}^2 = \frac{1}{M-1}\sum_{i=1}^{M-1}(x_i + \mu_t)^2$. Equation x demonstrates that the difference between the target neuron and other neurons is more noticeable at lower energies; therefore, the importance of the target neuron is expressed using $\frac{1}{e_t^*}$. At this point, E is used to define $\frac{1}{e_t^*}$ across all spatial dimensions and channels, followed by application of the sigmoid to $\frac{1}{E}$ As 3D weight and multiply it by the original input feature.

$$\tilde{x} = sigmoid\left(\frac{1}{E}\right) \Theta x \quad (3)$$

### 3.2.2. Mish Activation Function
Within the proposed YOLO-APD architecture, the conventional Rectified Linear Unit (ReLU) activation functions typically employed in convolutional blocks have been uniformly replaced by the Mish activation function. Mish is distinguished as a smooth, continuous, self-regularized, and non-monotonic activation, defined mathematically as:

$$f(x) = xtahh(softplus(x)) \quad (4)$$

One key strength of the Mish activation function lies in its continuous differentiability throughout an input domain, setting it apart from the piecewise-linear rectifier activation function (ReLU). Such smoothness encourages a more stable gradient flow in backpropagation, reducing potential optimization difficulties linked to singularities [54].

Moreover, Mish's unboundedness in the positive direction effectively counters saturation problems from large positive activations, which can impede training convergence. On the other hand, it remains bounded below (near ≈ -0.31), and its non-monotonic nature permits small negative inputs to yield negative outputs with non-zero gradients.

This behaviour is theorised to improve information propagation relative to ReLU, potentially avoiding the "dying ReLU" issue where neurons cease activity. Mish's intrinsic shape also offers a self-regularisation measure, which may contribute to better model generalization [44].

Consequently, implementing Mish across YOLO-APD's convolutional layers is intended to harness these beneficial traits for improved training stability, detection robustness, and overall model precision.

$$f(x) = xtahh(softplus(x)) = xtah(ln(1 + e^x)) \quad (5)$$

### 3.2.3. C3Ghost Module
To address the need to balance computational demands with feature representation capabilities, especially in extensive network designs, the C3Ghost module was introduced in this study. This module synergizes the structural benefits inherent in the Cross-Stage Partial Network (CSP) from the C3 module [28] with the efficient, lightweight convolution approach of GhostNet (Han et al., 2020).

A primary objective in designing the C3Ghost module was to lessen the computational load (FLOPs) and the parameter count relative to conventional convolutional blocks, while carefully maintaining the key discriminative features vital for the detection task.

The fundamental design of C3Ghost enhances the standard convolutions within the C3 module's bottleneck layers by incorporating Ghost convolutions, as illustrated in Figure 4.

The Ghost convolution operation first generates a small set of intrinsic feature maps through a standard convolution. Subsequently, these feature maps undergo less computationally demanding linear transformations—often depth-wise convolutions—to generate a more extensive array of "ghost" feature maps.

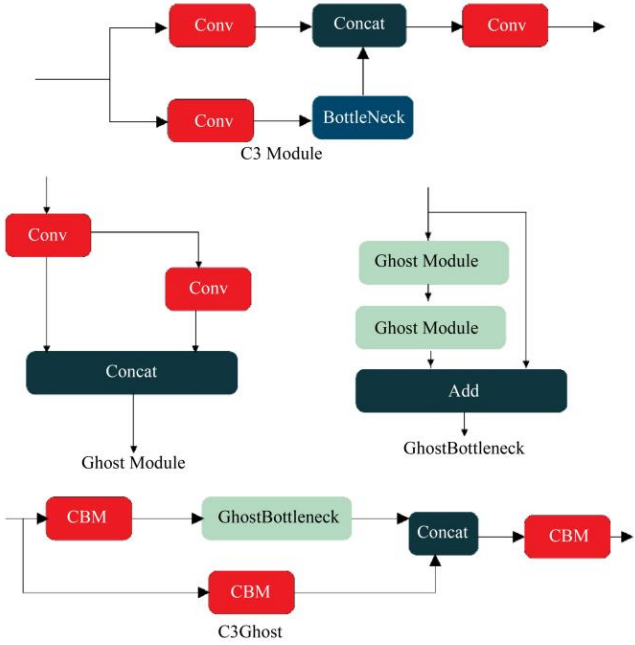

**Fig. 4 C3Ghost Module Structure**

These ghost maps effectively encapsulate redundant information with minimal processing overhead [41], substantially decreasing the computational effort required to produce the desired feature channel count. A further distinctive aspect of the C3Ghost module used in this study is the replacement of typical activation functions, such as ReLU or SiLU, with the Mish activation function.

The choice of Mish activation function was guided by its smoother gradient progression. Further, its capacity for stronger regularization compared with the other piecewise-linear alternatives made it an ideal candidate for improving the performance of the proposed model. These characteristics promote more stable training dynamics and offer pathways for improved model generalization [54].

Within the proposed model, the C3Ghost module acts as an essential underlying component. Central to its architecture is the effective generation of features stemming from Ghost convolutions. This is complemented by the incorporation of established Cross-Stage Partial (CSP) design principles from C3, coupled with the advantageous attributes of Mish activation.

Arranged this way, the C3Ghost module enables the construction of deep yet computationally manageable models, thereby realizing a practical trade-off between detection effectiveness and inference throughput.

### 3.2.4. SimSPPF Module

The YOLOv8 baseline algorithm improved the implementation of the Spatial Pyramid Pooling Fast (SPPF) module[55]—an innovation from earlier architectures like the YOLOv5 model and its predecessors. YOLOv8 optimization ensured computational efficiency by reducing the number of internal convolutional layers compared to earlier SPP variants [34].

Though it effectively reduced parameters and FLOPs, this simplification potentially led to a loss of fine-grained feature information, which affected detection precision for complex scenarios. To address this potential trade-off and enhance the feature representation capabilities of the SPPF block with a modest increase in computational cost, the authors propose replacing the SPPF module at the backbone of YOLOv8 with the SimSPPF module, as shown in Figure 2.

This modified module aims to improve detection accuracy, achieving a favourable balance between performance and efficiency. The architectural details and pseudocode for the proposed SimSPPF module are provided below in Fig. 5 and Table 1, respectively.

**Table 1. The Pseudocode of SimSPPF Module**

| SimSPPF Module | |
|---|---|
| Input: Feature maps of the road image with objects | |
| 1. | Define SimConv Module: |
| 2. | function Conv2d (cin, cout, k, s, g, bias=False) |
| 3. | function BatchNorm2d (cout, eps, momentum) |
| 4. | function Mish (BatchNorm2d(Conv2d(x))) |
| 5. | Define SimSPPF modules: |
| 6. | Set Parameters for MaxPool: kernel_size=5, stride (s) =1, padding (p) =2, α=0.1, eps=0.01 |
| 7. | Initialize Convolutional Layer c1=input_channels, c=intermediate_channels): |
| 8. | cv1 = SimConv(c1, c, k=1, stride=1, g=g_cv1) |
| 9. | cv2 = SimConv(4*c, c, k=3, stride=1, g=g_cv2) |
| 10. | Define the MaxPool2d function: |
| 11. | function MaxPool2d(features, kernel_size, stride, padding) |
| 12. | Forward operation: |
| 13. | x1 = cv1(input_features) |
| 14. | y1 = MaxPool2d (x1, 5, 1, 2) |
| 15. | y2 = MaxPool2d (y1, 5, 1, 2) |
| 16. | y3 = MaxPool2d (y2, 5, 1, 2) |
| 17. | concatenated_features = concatenate ((x1, y1, y2, y3), dim=1) |
| 18. | output_y = cv2(concatenated features |
| 19. | Return processed feature map output_y. |

First, the normal 2D convolution building block CBS (Conv + BatchNorm Silu) blocks are replaced with CBM (Conv BatchNorm Mish). The SimSPPF module processes the input feature maps from the preceding stages and then transforms these salient maps, using a SimConv layer (cv1). Following this, a pyramid of feature representations is constructed by applying MaxPool2d operations (kernel size 5, stride 1, padding 2) in series to the transformed features, effectively capturing contextual information at multiple effective receptive fields without altering spatial dimensions.

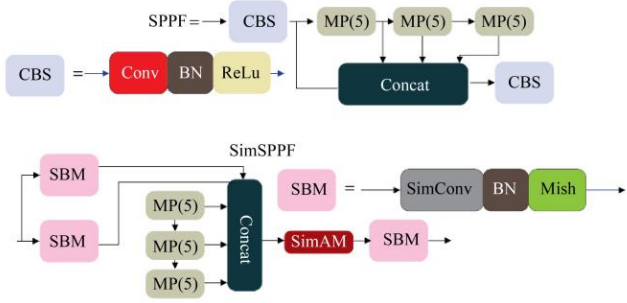

**Fig. 5 Structure of SimSPPF**

These parallel streams—the initially transformed features (x1) and their progressively pooled counterparts (y1, y2, and MaxPool2d(y2))—are then concatenated along the channel dimension. This concatenated feature tensor, rich in multi-scale information (resulting in 4c channels), is subsequently processed by a final SimConv layer (cv2), which judiciously fuses these diverse representations into a compact and powerful output feature map (typically c channels).

This architectural design allows the SimSPPF module to effectively aggregate contextual information, improving YOLO-APD's robustness to object scale variations while presenting a potentially more parameter-efficient architecture. The effectiveness of the SimSPPF module in enhancing detection performance relative to the standard SPPF is validated through comparative experiments detailed in the ablation studies section. The results presented therein demonstrate the contribution of SimSPPF to the overall improved model precision.

### 3.2.4. Intelligent Gather and Distribute (IGD) Module

The standard YOLOv8 architecture incorporates a neck structure founded on Feature Pyramid Network (FPN) [56] and Path Aggregation Network (PANet) [26] ideas. However, a key drawback of this conventional setup is its constrained information flow, where features mainly pass between adjacent levels. Such an architectural arrangement can diminish important information flow. Immediate layers can act as bottlenecks to the flow of fine-grained features as they propagate in a deeper network. This, in turn, may restrict the model's ability to effectively integrate global features across all scales, potentially affecting its performance on objects that need both fine-grained details and a wider contextual understanding.

To overcome this drawback, this work proposes replacing the existing YOLOv8 neck with a new Intelligent Gather-and-Distribute (IGD) module. This modification is intended to optimize the model for detecting objects in complex settings such as type S roads, other occluded scenes, and poorly lit settings. This approach draws upon the injection-multi and Sim4modules concepts, taking cues from the work of [57]. As illustrated in Figure 6, this architecture uses multiple branches to merge features from different scales through sequential top-down (for semantic enrichment) and bottom-up (for localization enhancement) pathways.

The IGD mechanism is engineered for a more comprehensive and parallelized interaction of features across scales, moving past simple pairwise fusion between adjacent levels. By allowing features from diverse levels to be gathered concurrently and then suitably redistributed, the model seeks to retain richer multi-scale information and boost the representational strength of the fused feature maps. It is posited that this improved fusion approach will lead to notable gains in detection accuracy, especially for anomalies that often display subtle or context-dependent characteristics within the proposed YOLO-APD architecture. The conventional Rectified Linear Unit (ReLU) activation functions typically employed in convolutional blocks have been uniformly replaced by the Mish activation function.

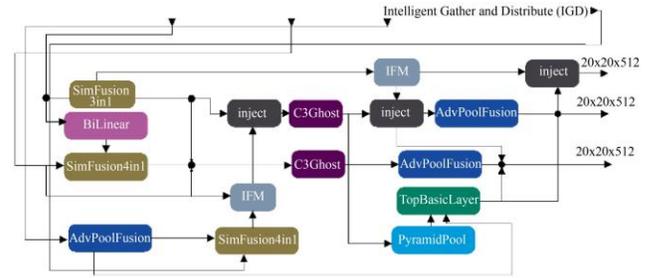

**Fig. 6 Intelligent Gather and Distribute Module**

### 3.3. Loss Function

The overall loss function of the YOLO detection algorithm is a weighted summation of three primary loss components. Classification loss ($L_{cls}$), bounding box regression loss ($L_{box}$) and distribution Focal Loss ($L_{DFL}$), typically expressed as:

$$L_{total} = \lambda_1 L_{cls} + \lambda_2 L_{box} + \lambda_3 L_{DFL} \qquad (6)$$

Where $\lambda_1$, $\lambda_2$, $\lambda_3$ are scaling hyperparameters. The classification loss component is denoted as ($L_{cls}$), leverages on Binary Cross Entropy with Logits (BCEWithLogitsLoss). This is applied to each predicted class score, effectively penalizing deviations from ground-truth labels. For the bounding box regression, ($L_{box}$), utilizes the complete intersection over Union (CioU) loss, which is expressed as.

$$L_{CIoU} = 1 - IoU + \frac{\rho^2(b, b^{gt})}{c^2} + \alpha v \qquad (7)$$

IOU is the intersection over Union between the predicted box and the ground truth box. $\alpha$ is given by $\frac{v}{(1-IoU)+v}$ and ν is given by $\frac{4}{\pi^2}(arctan\frac{w^{gt}}{h^{gt}} - arctan\frac{w}{h})^2$.

Complementing this, the Distribution Focal Loss (DFL), ($L_{DFL}$) models the continuous bounding box coordinates as a general distribution. For a constant target coordinate y that lies between two discrete, learnable bin edges $y_l$ and $y_r$, and the predicted probabilities $P(y_l)$ and $P(y_r$ For these edges, DFL is given by

$$L_{DFL} = -((y_r - y)\log(P(y_l)) + (y - y_r)\log(P(y_r))) \quad (8)$$

### 3.4. Integration with Dynamic Region of Interest (DROI)

In addition to the core network evaluation, this study presents a conceptual framework for integrating YOLO-APD into a system leveraging vehicle dynamics, as shown in Figure 7. Data from the vehicle's steering angle sensor could dynamically calculate a Region of Interest (ROI) corresponding to the immediate path ahead, particularly relevant during turns on roads like Type-S, as shown in Figure 8. Steering angle(0-30degrees) defines the critical region ahead of the moving AV, where potential hazards are determined.

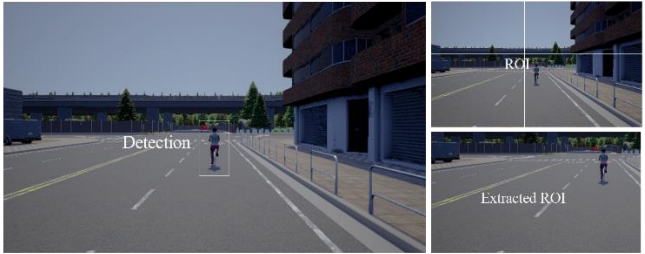

**Fig. 7 Dynamic Region Of Interest (DROI) Determination**

The critical region is then defined by.

$$w_c = W_0 + k_1.|\theta| + k_2.v \quad (9)$$

Where $W_0$ is the base width when driving straight, $\theta$ is the steering angle, v is the speed of the vehicle, while $k_1, k_2$ These tuning parameters balance the steering wheel influence, speed, and how much the region of interest expands. At straight driving (0° to 30°), When $v = 0$, $W_c$ reverts to $W_0$ to prevent unnecessary expansion. At the moderate curve (30° < $|\theta| \leq 60°$ The ROI expands laterally, influenced by the steering angle and vehicle speed as it anticipates the trajectory along the curve. For steering angles > 60°, the system extends the critical region laterally based on the predicted vehicle path to account for sharper turns. This adaptive scene processing based on the DROI could reduce the overall computation load.

This would involve modules for ROI calculation based on steering angle (θ) and speed (v). Other downstream components, such as the distance estimation module, Time-to-Collision (TTC) checks, and object tracking (e.g., using a Kalman filter), would be crucial for risk assessment. The overall coordination of these components would trigger warnings or control actions that would further improve the quality of the ADAS. This system-level integration remains a direction for future implementation[58].

## 4. Results and Discussion
### 4.1. Experimental Environment Configuration

**Table 2. Experiment configuration**

| Configuration | Details |
|---|---|
| Operating System | Windows 11 |
| CPU | Intel® Core(TM) i5-11400H |
| Running Memory | 16GB |
| GPU | Tesla P100-PCIE |
| Python | Version 3.8.0 |
| Pytorch | Version 1.8.0 |
| Compiling Software | Visual Studio Code 2023 |
| Unreal Engine | 4.27 |
| Simulator | Carla 0.9.15 |

Using the synthetic Carla dataset and the corresponding annotation files, a detection experiment is carried out, beginning with model training with parameters shown in Table 3, and final testing in the simulator. The hardware and software configuration environment of the simulation experiment is shown in Table 1 above.

To facilitate the analysis of the detection effect before and after the improvement, the performance comparison between different detection algorithms, four evaluation indexes are introduced to comprehensively describe the detection precision of the model, which includes accuracy metrics such as average precision (AP), mean average precision(mAP), mean F1 score(mF1), besides robustness analysis whose definitions are as follows:

$$AP = \int_0^1 P(R)dR \;, \; mAP = \frac{1}{N_{cls}}\sum_{i=1}^{N_{cls}} AP_i \quad (10)$$

$$mF1 = \frac{1}{N_{cls}}\sum_{i=1}^{N_{cls}} \frac{2.P_i.R_I}{P_I + R_I} \quad (11)$$

$$P = \frac{N_{TP}}{N_{TP} + N_{FP}}, \; R = \frac{N_{TP}}{N_{TP} + N_{FN}} \quad (12)$$

Where $N_{TP}$ indicates the quantity of accurately identified objects, $N_{TP} + N_{FP}$ represents the total objects detected by the model, and $N_{TP} + N_{FN}$ Is the number of true objects detected? $N_{Cls}$ Represents the total number of categories, which is 8 in this study. Moreover, the FLOPs (floating-point operations) and Params (parameters) are used to measure the object detection model's computational time complexity and parameter quantity, respectively.

### *4.2. Algorithm Comparison in CARLA Dataset*

To validate the performance of YOLO-APD, this paper conducted comparative experiments with other object detection models, including Faster RCNN, SSD, YOLOv5, YOLOv7, YOLOv8 and YOLOv9. Under the same dataset samples and training parameters configuration, these detection methods are evaluated by combining AP, mAP, F1, FLOPs, and Params evaluation indexes.

To quantify the capabilities of the evaluated models, a specific set of performance benchmarks was established. These benchmarks included AP@0.5, mAP, FPS, the mean F1 score (mF1), inference duration per image (TD), computational load (FLOPs), and the count of model parameters as summarised in Table 4. A consistent pattern emerges from these collected experimental figures: models invariably balance their achieved detection accuracy and the efficiency of their inference operations.

**Table 3. Training parameters configuration**

| Parameters | Details |
|---|---|
| Optimizer | AdamW. |
| Learning Rate | 0.01, |
| Batch Size | 16. |
| Epochs | 100 |
| Regularization | decay 0.0005, Mom 0.937. |
| Initialization | COCO pre-trained weights. |
| Augmentation | mosaic, horizontal flips |

While SSD achieved a relatively high FPS of 200, it had a significantly lower detection precision of mAP 60.5%. YOLOv5, a moderate performer, improves mAP to 66.1%, but its relatively low FPS (100) and higher inference latency (10 ms per image) hinder its suitability for real-time applications. In contrast, Faster R-CNN achieved a high detection accuracy (AP@0.5 of 90.1%, mAP of 62.5%). However, it recorded an impractical inference speed of 5 FPS. This performance bottleneck is due to its substantial computational overhead (169.82 GFLOPs), making it infeasible for deployment in real-time systems.

YOLOv7 and YOLOv8 showed notable improvements in detection performance and inference speed, with mAP scores of 67.7% and 71.5%, respectively, and real-time processing capabilities (~120 FPS). However, both models incurred significant computational costs: YOLOv7 required 157.1 GFLOPs. In contrast, YOLOv8, though slightly more efficient at 67.7 GFLOPs, is encumbered by a heavier parameter count (20.04M), which limits its scalability on resource-constrained embedded systems.

In this study, however, YOLO-APD outperformed all the other models. It achieved the highest mAP (77.7%) and an AP@0.5 (97.0%) compared to the different models. It is worth noting that, although its FPS (100) was marginally lower than that of SSD and YOLOv7, it remains within the acceptable range for real-time autonomous driving applications. YOLO-APD maintains a robust inference time of 10 ms per image, with its FLOPs (76.5G) and parameter count (24.16M) reflecting a balanced computational footprint, especially when compared to the more resource-intensive YOLOv7 and YOLOv8 models.

Despite a slight increase in memory usage relative to the SSD algorithm, the considerable gains in detection precision and real-time processing capability justify this trade-off. YOLO-APD's strong performance stems from a carefully engineered architecture. Key to its design is an advanced anchor-free detection system, paired with enhanced methods for feature fusion. This combination equips the model to more reliably navigate common difficulties in pedestrian detection, including occlusion, substantial changes in object scale, and environmental interference.

**Table 4. Algorithm comparative analysis**

| Model | AP@0.5(%) | mAP(%) | FPS | (mF1) | TD(ms/img) | FLOPs (G) | Params |
|---|---|---|---|---|---|---|---|
| SSD | 86.4 | 60.5 | 200 | 0.81 | 5 | 62.15 | 75.63 |
| Faster R-CNN | 90.1 | 62.5 | 5 | 0.83 | 200 | 169.82 | 4.13 |
| Yolov5 | 90.8 | 66.1 | 100 | 0.870 | 10 | 6.8 | 2.69 |
| Yolov7 | 93.2 | 67.7 | 120 | 0.882 | 8.3 | 157.1 | 2.50 |
| YOLOv8 | 93.1 | 71.5 | 120 | 0.910 | 8.3 | 67.7 | 20.04 |
| YOLO-APD | 97.0 | 77.7 | 100 | 0.944 | 10 | 76.5 | 24.16 |

#### *4.2.1. Analysis of the Confusion Matrixes*

The confusion matrixes were analysed to gain a deeper insight into the class-specific performance and potential misclassifications. The normalized confusion matrix (Figure 8) details the proportion of correct versus incorrect predictions per class, with raw counts presented in Figure 9.

For pedestrian detection, YOLO-APD achieved exceptional results with 100% accuracy (Precision and Recall of 1.00) on the test set for this crucial class (Figure 8). Such performance is a vital outcome concerning the safety aims of autonomous driving systems. Recognition accuracy for traffic light states (green, orange, red) and 'traffic_sign_90' was likewise very high, correctly classifying all instances. Nevertheless, certain misclassification patterns emerged. The confusion between motorbikes and bike classes was distinct; 7% of actual motorbikes were mistaken for bikes (Fig. 9), a consequence likely stemming from their visual resemblance.

Further minor confusion occurred between traffic_sign_30 and traffic_sign_60, where 7% of traffic_sign_30 instances were misidentified as traffic_sign_60. Some confusion was also observed; 14% of traffic_sign_30 instances were incorrectly assigned to background, pointing to occasional lapses in detecting these signs.

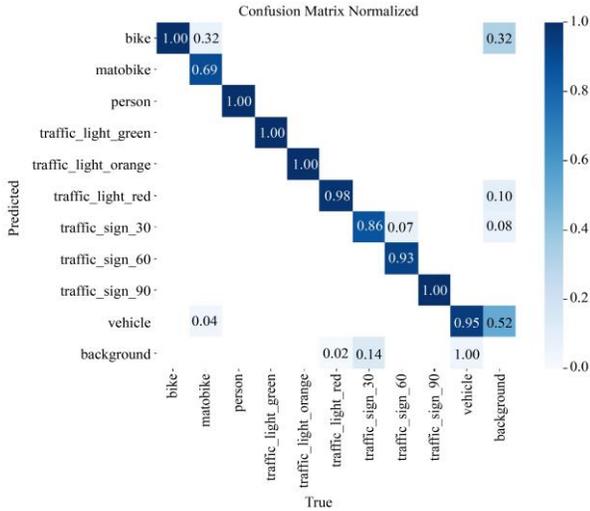

**Fig. 8 Normalized confusion matrix**

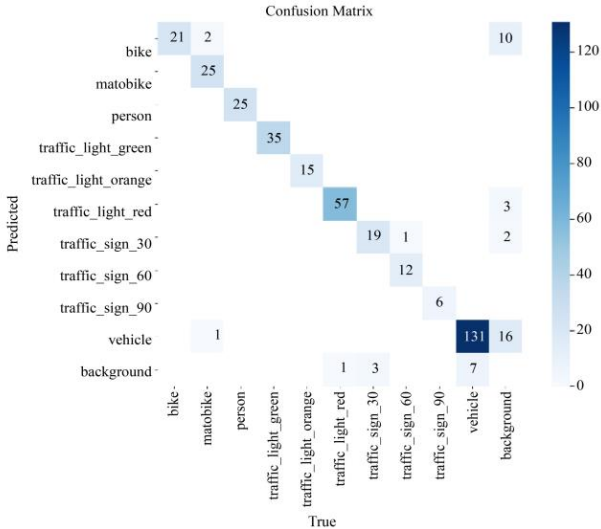

**Fig. 9 Raw confusion matrix**

A key observation from the normalized confusion matrix shown in Figure 8, rightmost column, further substantiated by the raw counts in Figure 9, is the high rate of false positives for vehicle and bike classes, which originated from background elements. From these observations, it is further noted that 52% of predictions categorized as vehicle and 32% of those identified as bike were background objects. This issue suggests a tendency for the model to misinterpret certain background patterns, which could lead to undesirable autonomous vehicle responses. Such a mishap could harm the overall safety of the AV and its occupants.

This challenge underscores an important direction for future model enhancements, potentially through focused negative data mining or by improving the distinctiveness of features against complex environmental backdrops. Even so, the model consistently attained 95% accuracy in correctly identifying genuine vehicle instances.

**Table 5. YOLO-APD Performance on KITTI and CARLA Datasets**

| Indexed Values | | KITTI | | CARLA | |
|---|---|---|---|---|---|
| | | AP | F1 | AP | F1 |
| YOLO-APD | Person | 0.965 | 0.817 | 0.981 | 0.9904 |
| | Car | 0.976 | 0.954 | 0.934 | 0.9249 |

### 4.3. KITTI and CARLA Dataset Comparison

A critical aspect of models trained on simulation is their ability to generalize to real-world data. To evaluate the experiment's reliability, the authors evaluated YOLO-APD, trained solely on the CARLA dataset developed for this study, on the KITTI road driving dataset (real-world dataset) for 'Person' and 'Car' classes. Table 5 above compares the performances. A marginal decrease in Average Precision (−0.016) and a substantial decline in the F1 score (−0.1734) were recorded on the KITTI dataset relative to CARLA.

This degradation is likely attributable to domain shift effects inherent between synthetic (CARLA) and real-world (KITTI) imagery, alongside the pronounced class imbalance within KITTI, where pedestrian instances are markedly underrepresented compared to vehicle classes. Such distributional disparities hinder the model's capacity to generalize effectively to minority classes under real-world conditions.

The study further revealed a contrast in the model's class performance on the KITTI and CARLA datasets. The real-world KITTI dataset recorded an increase of +0.042 in Average Precision and +0.0291 improvement in the F1 score compared to the CARLA dataset. This superior performance likely results from the higher fidelity and variability of vehicle instances in the real-world KITTI dataset, the later phenomenon leading to a superior generalization for the vehicular detection tasks.

Collectively, these findings highlight how adaptive object detection models are to the specifics of a given domain, such as the spread of data, the degree of environmental realism, and how frequently the presence of different classes significantly impacts performance. This, in turn, underscores the need for a robust domain adaptation by detection models.

To build a more resilient model that can adapt to domain shifts or skewed class representation as discussed above. It is critical to consider essential strategies. These include adjusting models with data from the target environment, developing methods to reconcile synthetic and real-world data characteristics, and creating data augmentation approaches to support underrepresented classes.

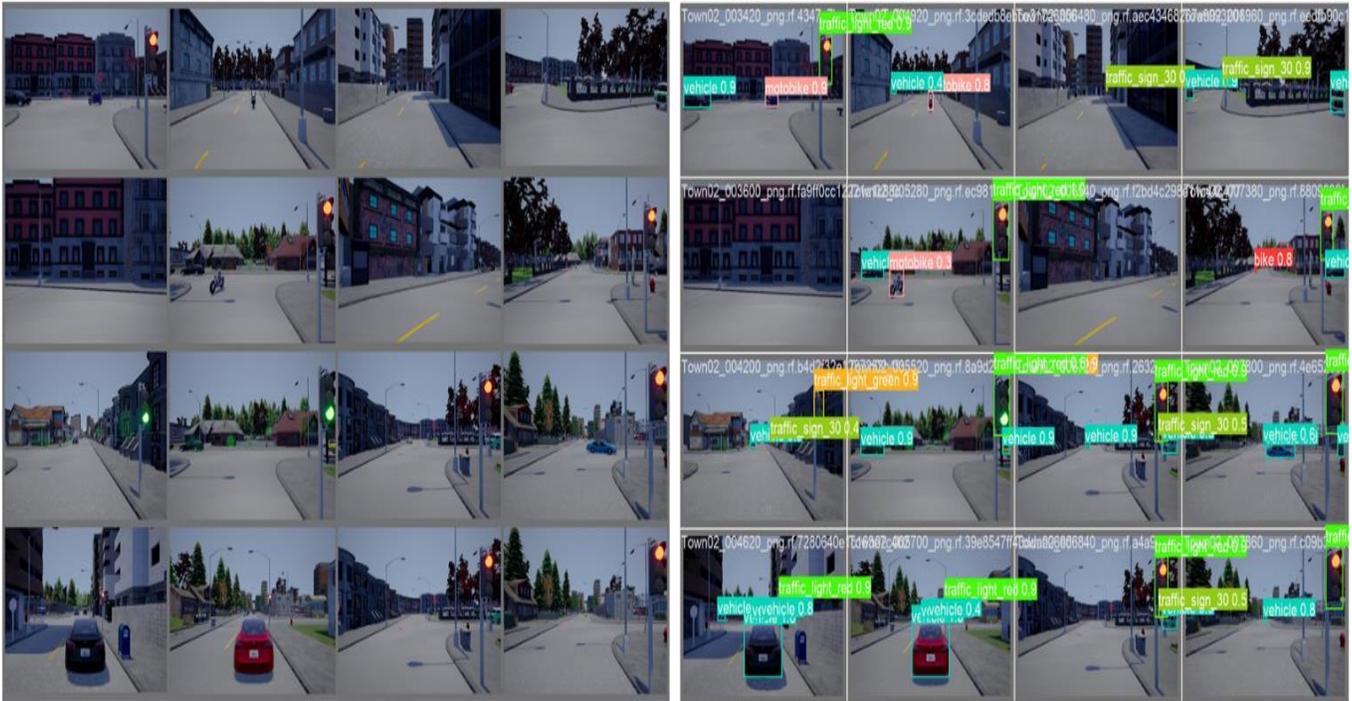

**Fig. 10 The left side shows the Scene before the algorithm detects, and the right side shows the Scene after detections by YOLO-APD**

*4.3.1. Graphical Illustrations of the YOLO-APD Performance*

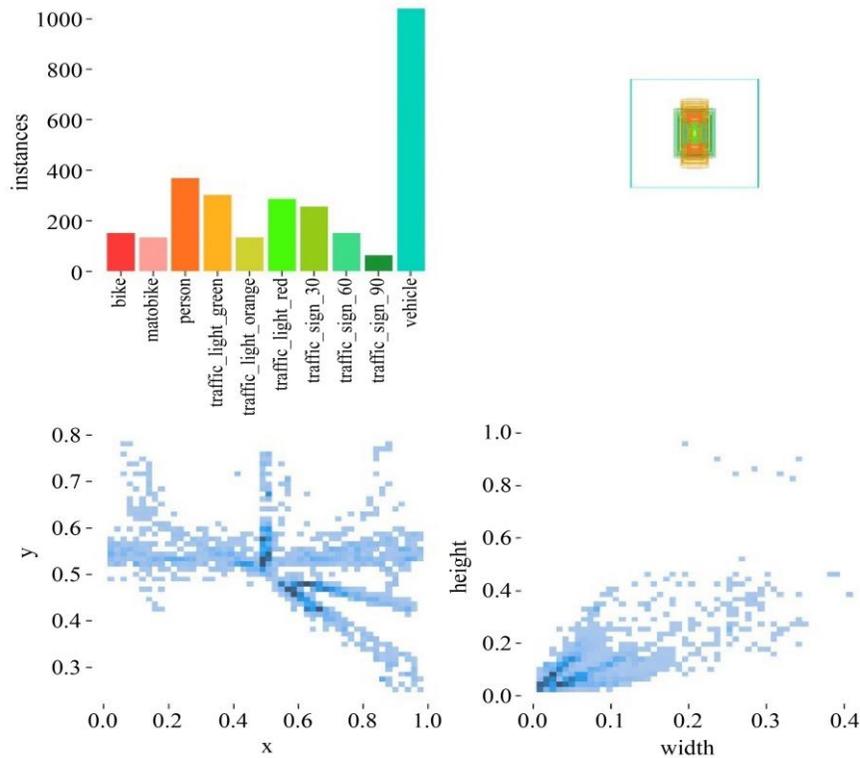

**Fig. 11 Dataset distribution illustrated the dataset's significant class imbalance, where 'vehicle' annotations predominate and others, such as 'traffic_sign_90,' are underrepresented. The bounding box overlay indicated predominantly centred object annotations, implying consistent labeling. An (x, y) heatmap of bounding box centres showed these concentrated near the image centre, mirroring the objects' spatial distribution. Furthermore, the width-height scatter plot demonstrated that most objects were relatively small, barring a few larger exceptions. This analysis highlighted that data augmentation or rebalancing strategies could be necessary for robust, all-class model performance.**

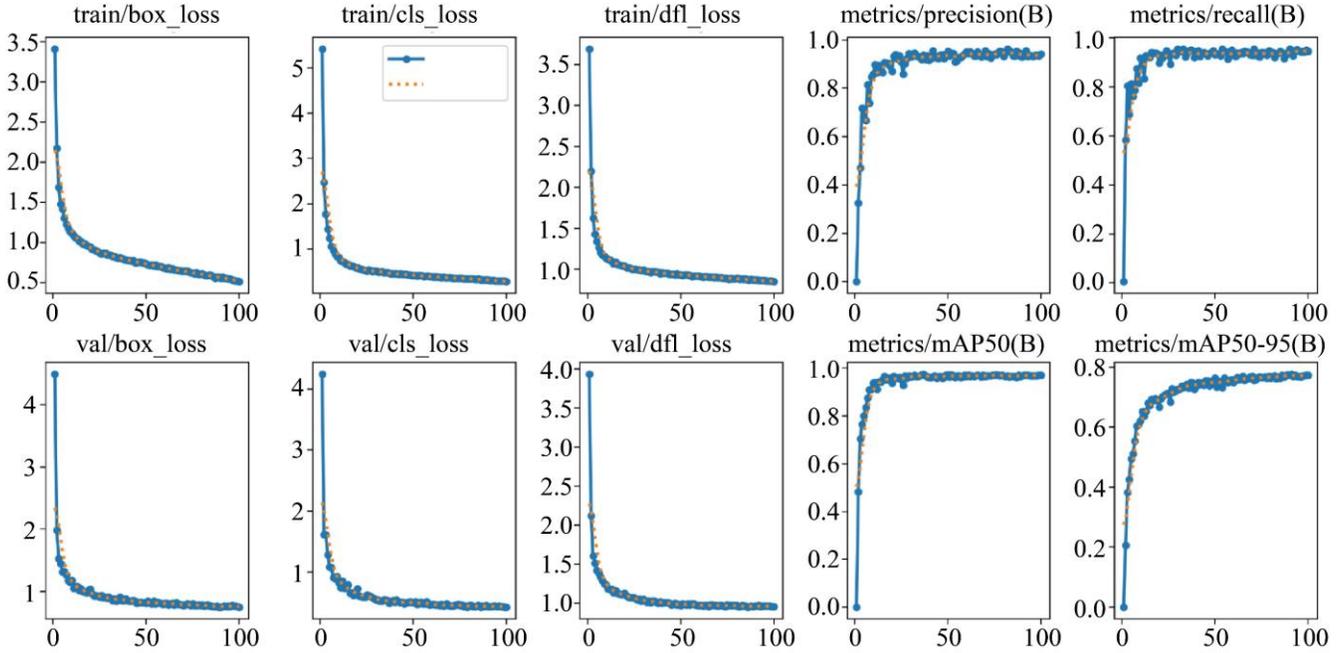

**Fig. 12 Training Losses for YOLO-APD**

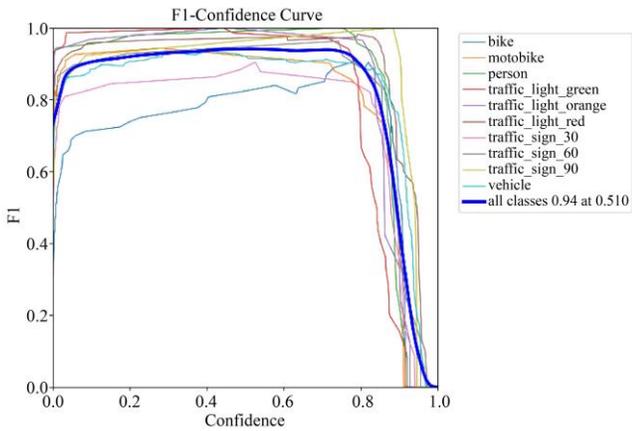

**Fig. 13 F1-Confidence Curve**

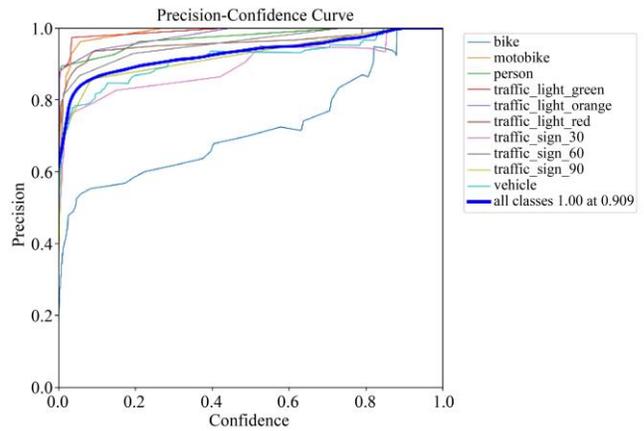

**Fig. 15 Graph of Precision-Confidence Curve**

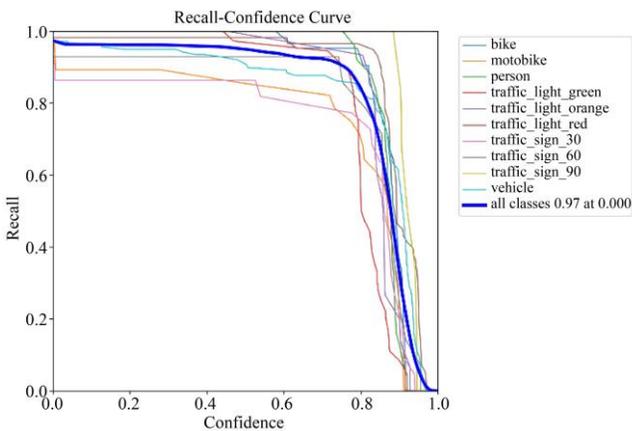

**Fig. 14 Graph of Recall-Confidence Curve**

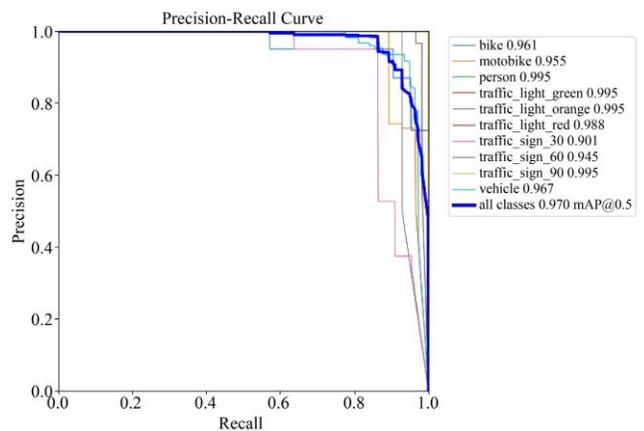

**Fig. 16 Graph of Precision-Recall Curve**

### 4.3.2. Graphical Illustrations of Comparative Performance

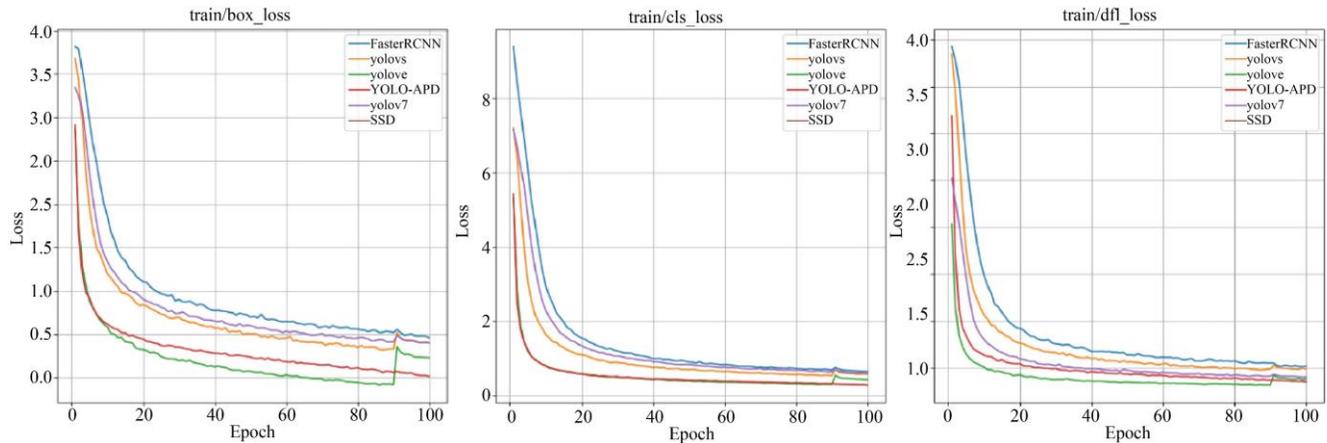

**Fig. 17 Individual training losses for the algorithms under comparison**

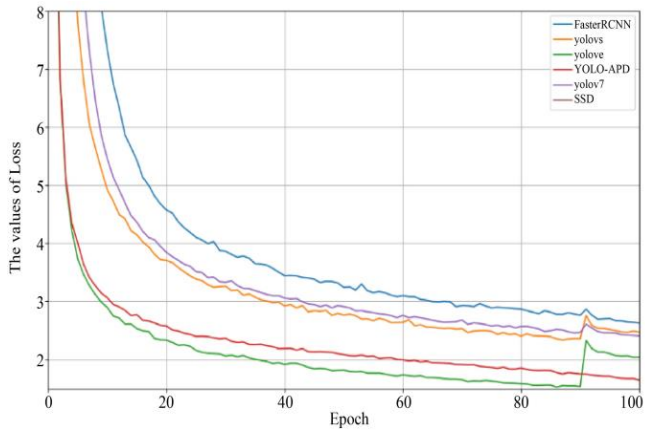

**Fig. 18 Graph of individual algorithm cumulative loss performance**

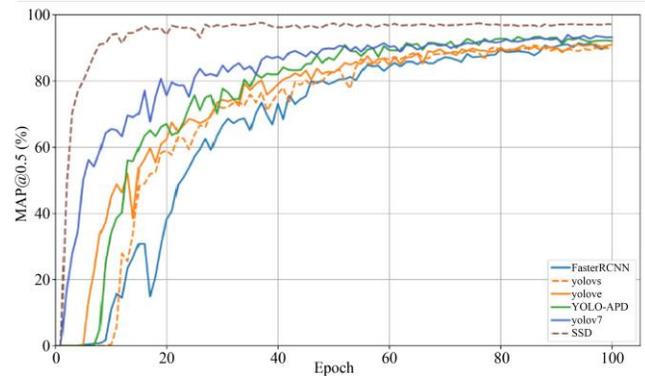

**Fig. 20 Graph of Comparative precision: YOLO-APD outperformed the other models, though there was a general uplift in performance for all the models**

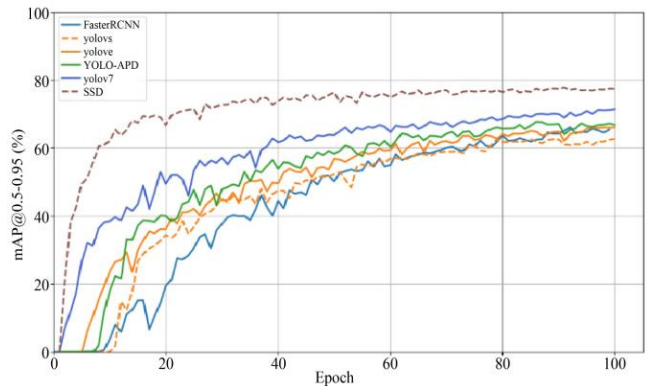

**Fig. 19 Graph of Comparative mAP for the different algorithms**

Focusing on mAP@0.5:0.95, YOLO-APD reached a peak accuracy approaching 80%. It also converged faster and demonstrated better overall precision than the other models. YOLOv8 followed, stabilizing near 70%, while YOLOv7 and YOLOv5 converged in the 66–68% bracket. Traditional architectures, namely Faster R-CNN and SSD, did not perform as well; SSD, notably, had issues with fluctuating performance and a slower learning trajectory.

### 4.4. Ablation Study

To evaluate the impact of YOLO-APD's specific architectural improvements, individually and combined, an ablation study was conducted using the CARLA dataset (detailed results appear in Table 6). The methodology began with the foundational YOLOv8 model. Subsequently, each distinct enhancement was incrementally introduced: initially, the SimSPPF module, then SimAM attention, followed by the head design drawing from Intelligent Gather-and-Distribute (IGD) concepts, and ultimately the Mish activation function.

The standard YOLOv8 model shown in Expr1 provided the initial performance benchmark, recording an AP@0.5 of 0.931 and an mAP@0.5 0.95 of 0.715. The first implemented modification (Exp2) centred on integrating the proposed SimSPPF module in place of the traditional SPPF. The overall effect of this isolated change was quite substantial, with AP@0.5 increasing to 0.952 (a relative gain of 2.1%) and mAP@0.5:0.95 rising to 0.729 (a 1.5% improvement). This immediate and significant performance enhancement suggests the inherent strengths of including the SimSPPF in the proposed model's architecture.

Table 6. Ablation Experiment

|       | SimSPPF | SimAM | GD | Mish | AP      | mAP     |
|-------|---------|-------|----|------|---------|---------|
| Expr1 |         |       |    |      | 0.93082 | 0.71453 |
| Exp2  | ✓       |       |    |      | 0.95188 | 0.72907 |
| Exp3  | ✓       | ✓     |    |      | 0.95341 | 0.73152 |
| Exp4  | ✓       | ✓     | ✓  |      | 0.95244 | 0.74564 |
| Exp5  | ✓       | ✓     | ✓  | ✓    | 0.97053 | 0.77444 |

SimSPPF's refined capability to discern and fuse multi-scale contextual features using an attention-augmented pooling strategy resulted in more robust and consistent detection outcomes. Subsequently, the integration of the parameter-free SimAM attention mechanism into the backbone, as shown in Exp3, led to a marginal but positive gain in performance: AP@0.5 rose to 0.953, while mAP@0.5:0.95 improved to 0.732 (a +0.2% absolute increase). Compared to the modular impact of SimSPPFs, these improvements were modest. However, they highlight SimAM's utility in selectively amplifying salient neuronal activations within the feature maps without adding learnable parameters. This modular inclusion contributed to improving YOLO-APD performance with minimal computational overhead.

The introduction of the IGD-inspired head (Exp4) yielded a notable trade-off. While the AP@0.5 metric experienced a marginal decrease to 0.952, a marked improvement was observed in the mAP@0.5:0.95 score, a key indicator of robustness across more stringent IoU criteria, which rose to 0.746 (a +1.4% absolute gain over Exp3). This divergence reveals the IGD head's crucial role in refining how well the model pinpoints objects and tells them apart. This enhanced localization capability came at the slight cost of fewer detections under the looser 0.5 IoU standard. However, for applications like autonomous driving, where high spatial fidelity is non-negotiable, this type of targeted improvement is highly beneficial.

Exp5 represents the final iteration of the YOLO-APD architecture. Mish activation was implemented across the entire network, producing the study's most significant jump in detection performance. The default SiLU activation functions were replaced strategically with Mish across several modules to improve feature extraction and performance. The performance of AP@0.5 improved to 0.971 (a +1.8% absolute rise over Exp4), and mAP@0.5:0.95, hitting 0.774 (a +2.8% absolute gain). This performance leap is likely attributed to Mish's smooth, non-monotonic nature. This smoothness promotes an improved gradient flow and a better optimization dynamic during training. The resulting model (YOLO-APD) achieves higher precision and demonstrates greater generalization across complex detection scenarios.

### 4.5. Discussion
This comprehensive experimental evaluation demonstrates the effectiveness of the proposed YOLO-APD network. A comparative analysis with state-of-the-art object detectors (Table 4) indicates that YOLO-APD performs competitively in the challenging pedestrian detection task within the simulated CARLA environment. The proposed model achieves a strong balance across key performance metrics, which includes a mean Average Precision (mAP@0.5:0.95) of 77.7%, a high mean F1 score (mF1) of 0.944, highlighting a recall performance which is critical for safety applications besides a real-time inference speed of 100 FPS, and moderate computational requirements (76.5 GFLOPs, 24.16 million parameters). YOLO-APD outperforms its direct baseline, YOLOv8m, across all reported accuracy metrics. The ablation study (Table 6) analyses YOLO-APD's performance based on modular improvements. It is shown that each architectural modification contributes to the overall gain in the model's efficiency. The integration of SimSPPF enhances multi-scale feature extraction, C3Ghost modules improve computational efficiency, SimAM (an attention mechanism without additional parameters) refines feature representations, the IG&D-inspired head improves feature fusion, and the Mish activation function supports better optimization.

The simulation to real-world evaluation using the KITTI dataset (as shown in Table 5) provides a practical insight into the generalization capabilities of the proposed YOLO-APD model. While the proposed model shows a relatively strong performance in the 'Car' category, it also highlights the challenges facing domain adaptation. The substantial drop in F1 score for the pedestrian category in the real-world KITTI dataset suggests that deploying models trained solely with synthetic datasets in simulated environments can be hindered by domain shift and class imbalance during real-world deployment. Addressing this gap may require strategies such as fine-tuning on a small set of target domain samples, domain randomization during simulation, domain-invariant feature learning, or synthetic-to-real domain translation using generative adversarial networks (GANs).

Despite this domain gap, strong performance shown by the model in the simulated CARLA environment and the real-world dataset suggests that YOLO-APD can learn robust, transferable features. In safety-centric applications like autonomous driving, failing to detect an obstacle, such as missing a pedestrian (a false negative), carries unacceptable risks. The CARLA dataset has demonstrated high recall in directly addressing this critical need. Separately, an intriguing possibility arises from utilizing steering angle data: such integration could permit a more fluid direction of computational power, offering a pathway to quicker inference processing.

## 5. Conclusion

This paper introduces YOLO-APD, an enhanced object detection network based on YOLOv8, specifically adapted for robust and efficient pedestrian detection in autonomous vehicles operating in complex road geometries such as Type-S roads. By integrating a novel SimSPPF module for multi-scale feature extraction, efficient C3Ghost blocks, the parameter-free SimAM attention mechanism, Mish activation, and a Gather-and-Distribute-inspired detection head, YOLO-APD achieves strong accuracy (77.7% mAP@0.5:0.95) and high pedestrian recall (>96%) on a challenging custom CARLA dataset. The proposed model effectively balances detection performance, speed (100 FPS), and computational efficiency when set against the baseline YOLOv8 and other established single-stage object detectors. While evaluation on the real-world KITTI dataset highlights challenges and the recognized imperative for smoother domain adaptation, this context also highlights that YOLO-APD offers a solid architectural foundation for real-world deployment. It marks a meaningful progress towards developing a reliable and adaptive pedestrian detection algorithm and a cost-effective RGB-based perception system for autonomous navigation in complex and dynamic environments.

A direction for future work is to address the simulation environment disparity to real-world performance by the proposed model, through targeted domain adaptation techniques. Further research could also involve improving the dynamic region-of-interest system based on vehicle dynamics, optimizing the model for embedded hardware deployment using techniques such as quantization, and exploring sensor fusion with complementary modalities.

## References

[1] M. Hassaballah et al., "Vehicle Detection and Tracking in Adverse Weather Using a Deep Learning Framework," *IEEE Transactions on Intelligent Transportation Systems*, vol. 22, no. 7, pp. 4230–4242, 2021. [CrossRef] [Google Scholar] [Publisher Link]

[2] Sumit Ranjan, and S. Senthamilarasu, *Applied Deep Learning and Computer Vision for Self-Driving Cars*, Packt Publishing, 2020. [Google Scholar] [Publisher Link]

[3] Alireza Razzaghi et al., "World Health Organization's Estimates of Death Related to Road Traffic Crashes and Their Discrepancy with Other Countries' National Report," *Journal of Injury and Violence Research*, vol. 12, no. 3, pp. 39-44, 2020. [CrossRef] [Google Scholar] [Publisher Link]

[4] Jin Qiu, Jian Liu, and Yunyi Shen, "Computer Vision Technology Based on Deep Learning," *2021 IEEE 2nd International Conference on Information Technology, Big Data and Artificial Intelligence ICIBA*, 2021. [CrossRef] [Google Scholar] [Publisher Link]

[5] Yi Cao, Yuning Wang, and Huijie Fan, "Improved YOLOv5s Network for Traffic Object Detection with Complex Road Scenes," *2023 IEEE 13th International Conference on CYBER Technology in Automation, Control, and Intelligent Systems (CYBER)*, 2023. [CrossRef] [Google Scholar] [Publisher Link]

[6] Johannes Deichmann, Autonomous Driving's Future: Convenient and Connected, McKinsey & Company, 2023. [Google Scholar] [Publisher Link]

[7] Gamze Akyol et al., "Deep Learning Based, Real-Time Object Detection for Autonomous Driving," *2020 28th Signal Processing and Communications Applications Conference (SIU)*, 2020. [CrossRef] [Google Scholar] [Publisher Link]

[8] Ali Ziryawulawo et al., "An Integrated Deep Learning-based Lane Departure Warning and Blind Spot Detection System: A Case Study for the Kayoola Buses," *2023 1st International Conference on the Advancements of Artificial Intelligence in African Context,* 2023. [CrossRef] [Google Scholar] [Publisher Link]

[9] Ramin Sahba, Amin Sahba, and Farshid Sahba, "Using a Combination of LiDAR, RADAR, and Image Data for 3D Object Detection in Autonomous Vehicles," *2020 11th IEEE Annual Information Technology, Electronics and Mobile Communication Conference (IEMCON)*, 2020. [CrossRef] [Google Scholar] [Publisher Link]

[10] Xiangmo Zhao et al., "Fusion of 3D LIDAR and Camera Data for Object Detection in Autonomous Vehicle Applications," *IEEE Sensors Journal*, vol. 20, no. 9, pp. 4901–4913, 2020. [CrossRef] [Google Scholar] [Publisher Link]

[11] Fan Bu et al., "Pedestrian Planar LiDAR Pose (PPLP) Network for Oriented Pedestrian Detection Based on Planar LiDAR and Monocular Images," *IEEE Robotics and Automation Letters*, vol. 5, no. 2, pp. 1626–1633, 2020. [CrossRef] [Google Scholar] [Publisher Link]

[12] Michal Uřičář et al., "VisibilityNet: Camera Visibility Detection and Image Restoration for Autonomous Driving," *Electronic Imaging*, vol. 32, pp. 79–1–79–8, 2020. [CrossRef] [Google Scholar] [Publisher Link]

[13] Richard Szeliski, *Computer Vision: Algorithms and Applications,* Second Edition, Springer, 2022. [Google Scholar] [Publisher Link]

[14] Hao Zhang, and Shuaijie Zhang, "Focaler-IoU: More Focused Intersection over Union Loss," *arXiv Preprint*, 2024. [CrossRef] [Google Scholar] [Publisher Link]

[15] Fred Hasselman, and Anna M.T. Bosman, "Studying Complex Adaptive Systems with Internal States: A Recurrence Network Approach to the Analysis of Multivariate Time-series Data Representing Self-reports of Human Experience," *Frontiers in Applied Mathematics and Statistics*, vol. 6, 2020. [CrossRef] [Google Scholar] [Publisher Link]

[16] Alex Krizhevsky, Ilya Sutskever, and Geoffrey E. Hinton, "ImageNet Classification with Deep Convolutional Neural Networks," *Communications of the ACM*, vol. 60, no. 6, pp. 84-90, 2017. [CrossRef] [Google Scholar] [Publisher Link]

[17] Kaiming He et al., "Mask R-CNN," *IEEE Transactions on Pattern Analysis and Machine Intelligence*, vol. 42, no. 2, pp. 386–397, 2020. [CrossRef] [Google Scholar] [Publisher Link]

[18] Ross Girshick, "Fast R-CNN," *2015 IEEE International Conference on Computer Vision (ICCV)*, 2015. [CrossRef] [Google Scholar] [Publisher Link]

[19] Shaoqing Ren et al., "Faster R-CNN: Towards Real-Time Object Detection with Region Proposal Networks," *IEEE Transactions on Pattern Analysis and Machine Intelligence*, vol. 39, no. 6, pp. 1137–1149, 2017. [CrossRef] [Google Scholar] [Publisher Link]

[20] Bilel Tarchoun et al., "Deep CNN-based Pedestrian Detection for Intelligent Infrastructure," *2020 5th International Conference on Advanced Technologies for Signal and Image Processing (ATSIP)*, 2020. [CrossRef] [Google Scholar] [Publisher Link]

[21] Yunchuan Wu, Cheng Chen, and Bo Wang, "Pedestrian Detection Based on Improved SSD Object Detection Algorithm," *2022 International Conference on Networking and Network Applications (NaNA)*, 2022. [CrossRef] [Google Scholar] [Publisher Link]

[22] Wei Liu et al., "SSD: Single Shot MultiBox Detector," *Computer Vision-ECCV 2016*, pp. 21–37, 2016. [CrossRef] [Google Scholar] [Publisher Link]

[23] Joseph Redmon et al., "You Only Look Once: Unified, Real-Time Object Detection," *2016 IEEE Conference on Computer Vision and Pattern Recognition*, 2016. [CrossRef] [Google Scholar] [Publisher Link]

[24] Juan Terven, Diana-Margarita Cordova-Esparza, and Julio-Alejandro Romero-Gonzalez, "A Comprehensive Review of YOLO: From YOLOv1 to YOLOv8 and Beyond," *Machine Learning and Knowledge Extraction*, vol. 5, no. 4, pp. 1680-1716, 2023. [CrossRef] [Google Scholar] [Publisher Link]

[25] Joseph Redmon, and Ali Farhadi, "YOLOv3: An Incremental Improvement," *arXiv Preprint*, 2018. [CrossRef] [Google Scholar] [Publisher Link]

[26] Alexey Bochkovskiy, Chien-Yao Wang, and Hong-Yuan Mark Liao, "YOLOv4: Optimal Speed and Accuracy of Object Detection," *arXiv Preprint*, 2020. [CrossRef] [Google Scholar] [Publisher Link]

[27] G. Jocher and others, "YOLOv5 by Ultralytics. 2020," 2023. [Google Scholar]

[28] Rahima Khanam, and Muhammad Hussain, "What is YOLOv5: A Deep Look into the Internal Features of the Popular Object Detector," *arXiv Preprint*, 2024. [CrossRef] [Google Scholar] [Publisher Link]

[29] Chuyi Li et al., "YOLOv6 v3.0: A Full-Scale Reloading," *arXiv Preprint*, 2023. [CrossRef] [Google Scholar] [Publisher Link]

[30] Chien-Yao Wang, Alexey Bochkovskiy, and Hong-Yuan Mark Liao, "YOLOv7: Trainable Bag-of-freebies Sets New State-of-the-art for Real-time Object Detectors," *2023 IEEE/CVF Conference on Computer Vision and Pattern Recognition*, 2023. [CrossRef] [Google Scholar] [Publisher Link]

[31] Muhammad Yaseen, "What is YOLOv8: An In-Depth Exploration of the Internal Features of the Next-Generation Object Detector," *arXiv Preprint*, 2024. [CrossRef] [Google Scholar] [Publisher Link]

[32] Fatma Betul Kara Ardaç, and Pakize Erdogmuş, "Car Object Detection: Comparative Analysis of YOLOv9 and YOLOv10 Models," *2024 Innovations in Intelligent Systems and Applications Conference (ASYU)*, 2024. [CrossRef] [Google Scholar] [Publisher Link]

[33] Priyanto Hidayatullah et al., "YOLOv8 to YOLO11: A Comprehensive Architecture In-depth Comparative Review," *arXiv preprint*, 2025. [CrossRef] [Google Scholar] [Publisher Link]

[34] Rejin Varghese, and M. Sambath, "YOLOv8: A Novel Object Detection Algorithm with Enhanced Performance and Robustness," *2024 International Conference on Advances in Data Engineering and Intelligent Computing Systems, ADICS 2024*, 2024. [CrossRef] [Google Scholar] [Publisher Link]

[35] Jie Hu et al., "Squeeze-and-Excitation Networks," *IEEE Transactions on Pattern Analysis and Machine Intelligence*, vol. 42, no. 8, pp. 2011–2023, Sep. 2018. [CrossRef] [Google Scholar] [Publisher Link]

[36] Qilong Wang et al., "ECA-Net: Efficient Channel Attention for Deep Convolutional Neural Networks," *Proceedings of the IEEE Computer Society Conference on Computer Vision and Pattern Recognition*, 2020. [CrossRef] [Google Scholar] [Publisher Link]

[37] Qibin Hou, Daquan Zhou, and Jiashi Feng, "Coordinate Attention for Efficient Mobile Network Design," *2021 IEEE/CVF Conference on Computer Vision and Pattern Recognition*, 2021. [CrossRef] [Google Scholar] [Publisher Link]

[38] Lingxiao Yang et al., "SimAM: A Simple, Parameter-Free Attention Module for Convolutional Neural Networks," *Proceedings of the 38th International Conference on Machine Learning*, pp. 11863–11874, 2021. [Google Scholar] [Publisher Link]

[39] Wei Li et al., "Object Detection based on an Adaptive Attention Mechanism," *Scientific Reports,* 2020. [CrossRef] [Google Scholar] [Publisher Link]

[40] Guoxin Shen, Xuerong Li, and Yi Wei, "Improved Algorithm for Pedestrian Detection of Lane Line based on YOLOv5s Model," *2022 IEEE 6th Advanced Information Technology, Electronic and Automation Control Conference (IAEAC)*, 2022. [CrossRef] [Google Scholar] [Publisher Link]

[41] Kai Han et al., "GhostNet: More Features from Cheap Operations," *2020 IEEE/CVF Conference on Computer Vision and Pattern Recognition*, pp. 1577-1586, 2020. [CrossRef] [Google Scholar] [Publisher Link]

[42] Shu Liu et al., "Path Aggregation Network for Instance Segmentation," *2020 IEEE/CVF Conference on Computer Vision and Pattern Recognition*, 2018. [CrossRef] [Google Scholar] [Publisher Link]

[43] Mingxing Tan, Ruoming Pang, and Quoc V. Le, "EfficientDet: Scalable and Efficient Object Detection," *2020 IEEE/CVF Conference on Computer Vision and Pattern Recognition*, 2020. [CrossRef] [Google Scholar] [Publisher Link]

[44] Diganta Misra, "Mish: A Self Regularized Non-Monotonic Activation Function," *arXiv preprint*, 2019. [CrossRef] [Google Scholar] [Publisher Link]

[45] Yinpeng Chen et al., "Dynamic Convolution: Attention Over Convolution Kernels," *2020 IEEE/CVF Conference on Computer Vision and Pattern Recognition*, 2020. [CrossRef] [Google Scholar] [Publisher Link]

[46] Chao Li, Aojun Zhou, and Anbang Yao, "Omni-Dimensional Dynamic Convolution," *arXiv Prepring*, 2022. [CrossRef] [Google Scholar] [Publisher Link]

[47] Lile Huo et al., "Overview of Pedestrian Detection based on Infrared Image," *2022 41st Chinese Control Conference (CCC)*, 2022. [CrossRef] [Google Scholar] [Publisher Link]

[48] Yu Song et al., "Full-Time Infrared Feature Pedestrian Detection Based on CSP Network," *2020 International Conference on Intelligent Transportation, Big Data & Smart City (ICITBS)*, 2020. [CrossRef] [Google Scholar] [Publisher Link]

[49] Timothe Verstraete, and Naveed Muhammad, "Pedestrian Collision Avoidance in Autonomous Vehicles: A Review," *Computers,* vol. 13, no. 3, p. 78, 2024. [CrossRef] [Google Scholar] [Publisher Link]

[50] Joel Janai et al., "Computer Vision for Autonomous Vehicles: Problems, Datasets and State of the Art," *Foundations and Trends® in Computer Graphics and Vision*, vol. 12, no. 1–3, pp. 1–308, 2020. [CrossRef] [Google Scholar] [Publisher Link]

[51] Holger Caesar et al., "nuScenes: A Multimodal Dataset for Autonomous Driving," *Proceedings of the 2020 IEEE/CVF Conference on Computer Vision and Pattern Recognition*, pp. 11621-11631, 2020. [Google Scholar] [Publisher Link]

[52] Gemb Kaljavesi et al., "CARLA-Autoware-Bridge: Facilitating Autonomous Driving Research with a Unified Framework for Simulation and Module Development," *2024 IEEE Intelligent Vehicles Symposium*, 2024. [CrossRef] [Google Scholar] [Publisher Link]

[53] Peiyu Yang et al., "A Part-Aware Multi-Scale Fully Convolutional Network for Pedestrian Detection," *IEEE Transactions on Intelligent Transportation Systems*, vol. 22, no. 2, pp. 1125–1137, 2021. [CrossRef] [Google Scholar] [Publisher Link]

[54] Rupshali Dasgupta, Yuvraj Sinha Chowdhury, and Sarita Nanda, "Performance Comparison of Benchmark Activation Function ReLU, Swish and Mish for Facial Mask Detection Using Convolutional Neural Network," *Intelligent Systems*, pp. 355–367, 2021. [CrossRef] [Google Scholar] [Publisher Link]

[55] Kaiming He et al., "Spatial Pyramid Pooling in Deep Convolutional Networks for Visual Recognition," *IEEE Transactions on Pattern Analysis and Machine Learning*, vol. 37, no. 9, pp. 1904-1916, 2015. [CrossRef] [Google Scholar] [Publisher Link]

[56] Yichen Zhang et al., "A New Architecture of Feature Pyramid Network for Object Detection," *2020 IEEE 6th International Conference on Computer and Communications (ICCC)*, 2020. [CrossRef] [Google Scholar] [Publisher Link]

[57] Chengcheng Wang et al., "Gold-YOLO: Efficient Object Detector via Gather-and-Distribute Mechanism," Advances in Neural Information Processing Systems, 2023. [Google Scholar] [Publisher Link]

[58] Ken Arioka, and Yuichi Sawada, "Improved Kalman Filter and Matching Strategy for Multi-Object Tracking System," *2023 62nd Annual Conference of the Society of Instrument and Control Engineers (SICE)*, 2023. [CrossRef] [Google Scholar] [Publisher Link]